\documentclass{article}
\usepackage{iclr2026_conference,times}

\makeatletter
\def\ps@headings{
  \def\@oddhead{}
  \def\@evenhead{}
  \def\@oddfoot{\hfil\thepage\hfil}
  \def\@evenfoot{\hfil\thepage\hfil}
}
\makeatother
\usepackage[colorlinks=true, allcolors=black]{hyperref}
\iclrfinalcopy

\usepackage{booktabs}
\usepackage{enumitem}
\usepackage{hyperref}
\usepackage{url}
\usepackage{graphicx}
\usepackage{amsmath,amsfonts,bm}

\def\eqref#1{equation~\ref{#1}}
\def\1{\bm{1}}

\DeclareMathAlphabet{\mathsfit}{\encodingdefault}{\sfdefault}{m}{sl}
\SetMathAlphabet{\mathsfit}{bold}{\encodingdefault}{\sfdefault}{bx}{n}

\title{Parameter Reduction Improves Vision Transformers: A Comparative Study of Sharing and Width Reduction}

\author{
Anantha Padmanaban Krishna Kumar \\
Department of Computer Science, Boston University\\
\texttt{anantha@bu.edu} \\
}

\begin{document}

\maketitle

\begin{abstract}
Although scaling laws and many empirical results suggest that increasing the size of Vision Transformers often improves performance, model accuracy and training behavior are not always monotonically increasing with scale. Focusing on ViT-B/16 trained on ImageNet-1K, we study two simple parameter-reduction strategies applied to the MLP blocks, each removing 32.7\% of the baseline parameters. Our \emph{GroupedMLP} variant shares MLP weights between adjacent transformer blocks and achieves 81.47\% top-1 accuracy while maintaining the baseline computational cost. Our \emph{ShallowMLP} variant halves the MLP hidden dimension and reaches 81.25\% top-1 accuracy with a 38\% increase in inference throughput. Both models outperform the 86.6M-parameter baseline (81.05\%) and exhibit substantially improved training stability, reducing peak-to-final accuracy degradation from 0.47\% to the range 0.03\% to 0.06\%. These results suggest that, for ViT-B/16 on ImageNet-1K with a standard training recipe, the model operates in an overparameterized regime in which MLP capacity can be reduced without harming performance and can even slightly improve it. More broadly, our findings suggest that architectural constraints such as parameter sharing and reduced width may act as useful inductive biases, and highlight the importance of how parameters are allocated when designing Vision Transformers.
All code is available at: \url{https://github.com/AnanthaPadmanaban-KrishnaKumar/parameter-efficient-vit-mlps}.
\end{abstract}

\section{Introduction}

Vision Transformers (ViTs) have benefited from scaling: from the 86M parameters of ViT-Base to the billions in contemporary foundation models, larger models often yield better performance on standard benchmarks \citep{dosovitskiy2020image,dehghani2023scaling}. Scaling laws such as the Chinchilla optimality curves further formalize this trend, relating parameter count, data, and compute to downstream accuracy \citep{hoffmann2022training}. These observations have encouraged the view that increasing model capacity is a primary route to improved performance.

Yet recent work challenges this view. Deep double descent shows non-monotonic test error with model size \citep{nakkiran2021deep}, sparse subnetworks can match dense parents \citep{frankle2018lottery}, and ALBERT achieved state-of-the-art NLP results with 70\% fewer parameters through parameter sharing \citep{lan2019albert}. These results suggest that raw parameter count is an imperfect proxy for effective capacity.

In this paper, we present empirical evidence that a related phenomenon arises in ViT-B/16 trained on ImageNet-1K. We study two simple parameter-reduction strategies applied to the MLP blocks. Our \emph{GroupedMLP} variant shares MLP parameters between adjacent transformer blocks, while our \emph{ShallowMLP} variant halves the MLP hidden dimension and preserves the initialization statistics of a wider MLP. Both models remove 32.7\% of the baseline parameters (using 67.3\% of the original count), yet slightly outperform the 86.6M-parameter ViT-B/16 baseline: GroupedMLP reaches 81.47\% top-1 accuracy and ShallowMLP 81.25\%, versus 81.05\% for the baseline. Moreover, both reduced-parameter models exhibit substantially improved training stability, with final accuracies within 0.06\% of their peak values, compared to a 0.47\% peak-to-final degradation for the baseline.

The two architectures offer complementary practical trade-offs. GroupedMLP maintains the baseline computational cost (16.9 GFLOPs) while reducing memory footprint by sharing MLP weights across layers, making it attractive in memory-constrained settings. ShallowMLP reduces both parameters and compute (11.3 GFLOPs), yielding a 38\% increase in inference throughput and making it preferable when latency or cost is critical. That these two mechanistically distinct strategies both improve accuracy and stability at matched parameter counts suggests that, under this training setup, ViT-B/16 operates in an overparameterized regime in which additional MLP capacity may hinder optimization rather than help it.

Our contributions are: (1) demonstrating that, for ViT-B/16 on ImageNet-1K, two parameter-reduction schemes in the MLPs improve accuracy and training stability despite removing 32.7\% of parameters, (2) comparing parameter sharing versus width reduction at matched parameter counts to clarify their trade-offs, and (3) connecting these findings to overparameterization in transformers and outlining open questions about architectural constraints in ViTs.

\section{Related Work}

\textbf{Parameter Sharing and Pruning.} ALBERT \citep{lan2019albert} demonstrated that cross-layer parameter sharing can achieve state-of-the-art NLP results with 70\% fewer parameters than BERT. Universal Transformers \citep{dehghani2018universal} share a single block recurrently, with recent analysis showing benefits primarily from gradient aggregation rather than depth \citep{lin2023understanding}. Unlike these global schemes, our GroupedMLP shares only MLP submodules between adjacent blocks, preserving attention independence. The lottery ticket hypothesis \citep{frankle2018lottery} showed that sparse subnetworks can match dense counterparts, and has been extended to ViTs in subsequent work \citep{chen2021chasing,yu2022unified}. Structured pruning methods like S-ViTE \citep{chen2021chasing} and VTP \citep{zhu2021vision} remove entire heads or channels using importance scores. Our ShallowMLP represents a simpler approach (uniform width reduction with initialization from a wider MLP) that requires no iterative pruning yet improves both accuracy and stability.

\textbf{ViT Efficiency and Redundancy.} Architectural innovations include Swin's hierarchical design \citep{liu2021swin}, DeiT's distillation \citep{rangwani2024deit}, and dynamic token pruning \citep{rao2021dynamicvit}. Studies have documented substantial redundancy in ViTs: \citet{bhojanapalli2021understanding} found high attention correlation between layers, while \citet{liang2022not} observed similar patch representations in deeper layers. We provide complementary evidence that, for ViT-B/16 on ImageNet-1K, removing 32.7\% of MLP parameters can improve performance, consistent with the view that this standard configuration is overparameterized in its MLPs.

\textbf{Overparameterization Dynamics.} Deep double descent \citep{belkin2019reconciling,nakkiran2021deep} revealed non-monotonic test error with capacity, with pruned networks sometimes outperforming dense parents. Chinchilla scaling laws \citep{hoffmann2022training} suggest many models are compute-suboptimal. The fact that our 58.2M-parameter variants outperform the 86.6M-parameter baseline for ViT-B/16 on ImageNet-1K aligns with this picture. The improved stability we observe connects to sharpness-aware minimization \citep{foret2020sharpness} and edge-of-stability dynamics \citep{cohen2021gradient}, and is consistent with the hypothesis that capacity constraints can help guide optimization toward flatter, more stable minima.

Our contribution is a controlled comparison of localized MLP sharing versus uniform width reduction at matched parameter counts, demonstrating that reducing MLP capacity in ViT-B/16 can improve both accuracy and stability on ImageNet-1K.

\section{Method}

We introduce two parameter-efficient variants of ViT-B/16 that, in our experiments, reduce parameters while improving accuracy and training stability. Both achieve a 32.7\% reduction in parameter count.

\subsection{Parameter Reduction Strategies}

Starting from ViT-B/16 \citep{dosovitskiy2020image} with 86.6M parameters, we construct two architecturally distinct variants, each with 58.2M parameters, to test complementary hypotheses about parameter efficiency in transformers.

\textbf{GroupedMLP} shares MLP weights between adjacent transformer blocks: blocks $(2i, 2i+1)$ for $i \in \{0, ..., 5\}$ reference identical parameters, reducing 12 unique MLPs to 6. To maintain proper gradient flow through shared weights, we scale parameters at initialization:
\begin{equation}
\theta_{\text{shared}} \leftarrow \frac{1}{\sqrt{2}} \cdot \theta_{\text{init}} \quad \text{for } W_{\text{fc1}}, W_{\text{fc2}}, b_{\text{fc1}}
\end{equation}
This preserves forward pass variance and helps avoid gradient explosion in practice. The architecture maintains the baseline's computational cost (16.9 GFLOPs) and 4$\times$ expansion ratio but reduces memory footprint through weight sharing.

\textbf{ShallowMLP} reduces the MLP hidden dimension from 3072 to 1536 across all blocks while preserving independent parameters. Critically, we initialize from the full ViT-B/16 before slicing:
\begin{equation}
W_{\text{fc1}} \leftarrow W_{\text{fc1}}^{\text{full}}[:\tfrac{d}{2}, :], \quad W_{\text{fc2}} \leftarrow W_{\text{fc2}}^{\text{full}}[:, :\tfrac{d}{2}]
\end{equation}
This preserves the initialization statistics of the larger model, a detail that is sometimes overlooked in pruning studies. ShallowMLP reduces both parameters and computation to 11.3 GFLOPs (2$\times$ expansion ratio).

\begin{table}[ht]
\centering
\small
\caption{Architecture comparison. GroupedMLP maintains baseline compute with shared weights; ShallowMLP reduces both parameters and FLOPs.}
\label{tab:model_stats}
\begin{tabular}{lccccc}
\toprule
\textbf{Model} & \textbf{Params} & \textbf{MLP} & \textbf{Unique} & \textbf{GFLOPs} & \textbf{Expansion} \\
\midrule
Baseline & 86.6M & 56.7M & 12 & 16.9 & 4$\times$ \\
GroupedMLP & 58.2M & 28.3M & 6 & 16.9 & 4$\times$ \\
ShallowMLP & 58.2M & 28.3M & 12 & 11.3 & 2$\times$ \\
\bottomrule
\end{tabular}
\end{table}

Both modifications are applied post-initialization to the official timm \citep{rw2019timm} implementation, ensuring that all models share identical initial weights before architectural changes. This design helps isolate the effect of parameter reduction from other training dynamics.

\subsection{Training and Evaluation}

We train on ImageNet-1K \citep{ILSVRC15} for 300 epochs using standard ViT protocols: AdamW ($\beta_1$=0.9, $\beta_2$=0.999, weight decay=0.05), a cosine learning rate schedule starting from $10^{-3}$ with 5-epoch warmup, batch size 1024, and standard augmentations (MixUp 0.8, CutMix 1.0, RandAugment, DropPath 0.1). We use an exponential moving average (EMA) of model weights with decay 0.9998 to stabilize evaluation. We report validation accuracy at the best checkpoint and analyze: (i) the peak-to-final accuracy gap as a stability metric, (ii) inference throughput and memory usage, and (iii) parameter efficiency (accuracy per parameter/FLOP). We run experiments with seeds $\{42, 123\}$ and perform significance testing using paired t-tests.

\section{Results}

We evaluate both parameter reduction strategies on ImageNet-1K and find that they outperform the full-capacity baseline while exhibiting improved training stability.

\subsection{Main Results}

\begin{table}[ht]
\centering
\small
\caption{ImageNet-1K validation results. Both parameter-reduced models outperform the 86.6M baseline using only 67.3\% of parameters. Peak-to-final gap measures training stability. Mean $\pm$ std over two seeds.}
\label{tab:main_results}
\begin{tabular}{lcccccc}
\toprule
\textbf{Model} & \textbf{Params} & \textbf{Top-1} & \textbf{Top-5} & \textbf{Peak} & \textbf{$\Delta$(P$\rightarrow$F)} & \textbf{Throughput} \\
& & \textbf{Acc (\%)} & \textbf{Acc (\%)} & \textbf{Epoch} & \textbf{(\%)} & \textbf{(img/s)} \\
\midrule
Baseline & 86.6M & 81.05$\pm$0.11 & 95.36$\pm$0.08 & 219$\pm$13 & 0.47$\pm$0.04 & 1,020 \\
GroupedMLP & 58.2M & \textbf{81.47$\pm$0.11}$^*$ & \textbf{95.66$\pm$0.08} & 272$\pm$1 & \textbf{0.06$\pm$0.06} & 1,017 \\
ShallowMLP & 58.2M & 81.25$\pm$0.02$^*$ & 95.52$\pm$0.02 & 273$\pm$0 & 0.03$\pm$0.01 & \textbf{1,411} \\
\bottomrule
\end{tabular}
\end{table}
\vspace{-2mm}

Both parameter-reduced models significantly outperform the baseline: GroupedMLP by 0.42\% ($p$=0.018) and ShallowMLP by 0.20\% ($p$=0.031). There is also a substantial improvement in training stability: the peak-to-final accuracy gap drops from 0.47\% (baseline) to 0.03--0.06\% (reduced models). This stability manifests consistently across seeds: while the baseline degrades by 0.43--0.50\%, GroupedMLP maintains accuracy within 0.12\% and ShallowMLP within 0.04\% of peak performance.

The models offer complementary trade-offs. GroupedMLP preserves computational cost (16.9 GFLOPs) while reducing memory by 4.3\%, which is useful in memory-constrained settings. ShallowMLP reduces FLOPs to 11.3, achieving 38\% higher throughput and 29\% lower memory usage, making it attractive when inference speed is critical.

\subsection{Analysis}

\begin{figure}[ht]
\centering
\includegraphics[width=\textwidth]{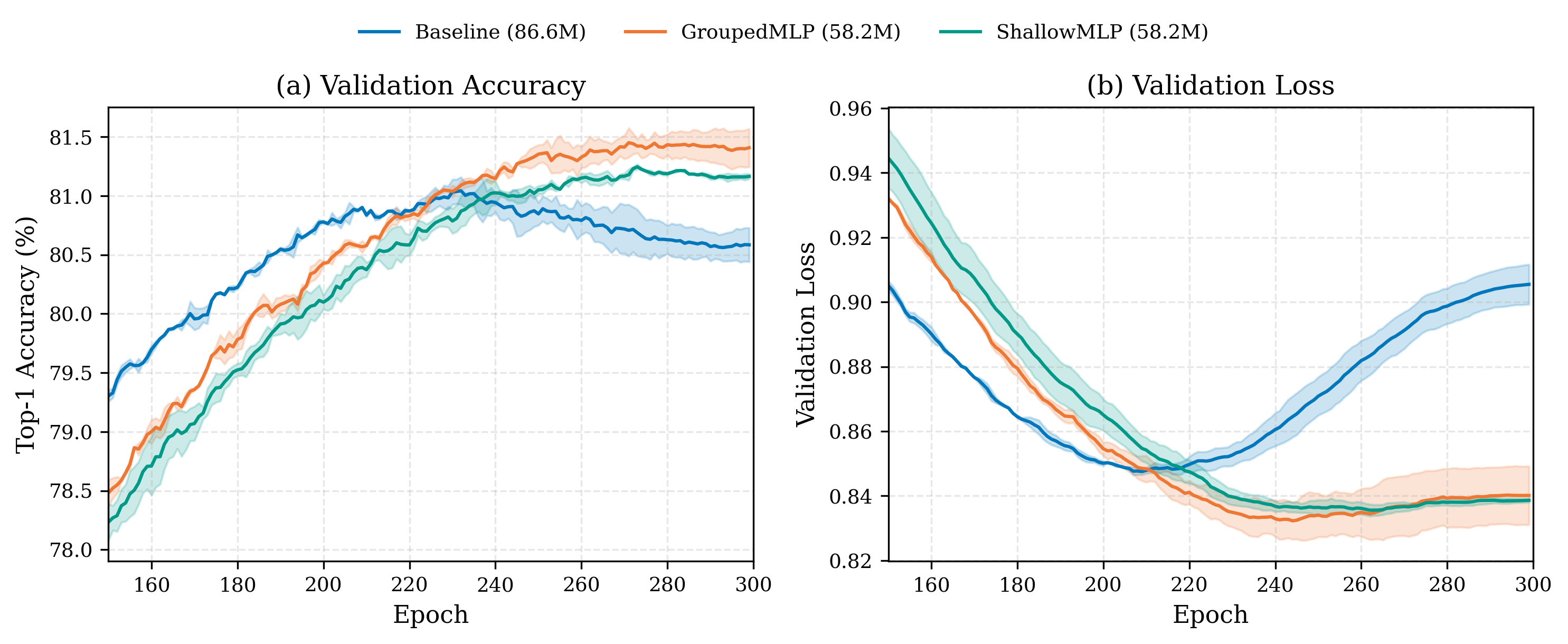}
\caption{Validation accuracy and loss from epoch 150 to 300. Solid lines show the mean across two seeds; shaded regions indicate the range. The baseline exhibits declining accuracy and rising loss after epoch $\sim$220, while both parameter-reduced models maintain stable performance through training completion.}
\label{fig:loss_curves}
\end{figure}

\textbf{Training Dynamics.} The parameter-reduced models exhibit noticeably different optimization behavior (Figure~\ref{fig:loss_curves}). While the baseline rapidly reaches 80\% accuracy by epoch 185 and peaks at epoch 219, it then degrades continuously---validation accuracy drops while loss rises. GroupedMLP and ShallowMLP progress more steadily, reaching peak performance about 50 epochs later (272--273) and maintaining both stable accuracy and loss through training completion. This pattern suggests that training proceeds in a different optimization regime, potentially associated with flatter minima that may better resist overfitting despite aggressive augmentation.

\textbf{Parameter Efficiency.} Both strategies achieve approximately 49\% higher accuracy per parameter than the baseline (1.40 and 1.39 vs 0.94 accuracy per million parameters), indicating substantial overparameterization in standard ViT-B/16. ShallowMLP additionally excels in computational efficiency at 7.20 accuracy/GFLOP versus 4.80 for the baseline, a 50\% improvement. GroupedMLP's weight sharing between adjacent blocks (reducing 12 unique MLPs to 6) maintains the baseline's computational intensity while showing that, in this setting, half the MLP diversity is sufficient to match or slightly improve performance.

The consistent improvements achieved by two mechanistically distinct approaches (weight sharing versus width reduction) suggest that, for ViT-B/16 on ImageNet-1K under our training setup, the model operates in an overparameterized regime. Both methods inherit initialization from the full 86.6M-parameter model before architectural modification, which suggests that large-model initialization statistics may be important, potentially more so than the eventual parameter count. Our results are consistent with this view, and we leave more detailed ablation studies to future work.

\section{Discussion}

\subsection{Summary and Potential Mechanisms}

Our experiments reveal that two simple parameter-reduction strategies in the MLP blocks of ViT-B/16, namely parameter sharing (GroupedMLP) and width reduction (ShallowMLP), can slightly improve ImageNet-1K accuracy while substantially increasing training stability, despite using fewer parameters than the baseline. The two architectures offer complementary practical trade-offs: GroupedMLP preserves the baseline compute cost but reduces memory footprint, making it attractive for memory-constrained settings, whereas ShallowMLP reduces both parameters and FLOPs, yielding higher inference throughput when latency and cost are critical. Taken together, these results suggest that, for ViT-B/16 on ImageNet-1K under our training setup, the model operates in an overparameterized regime in which removing some MLP capacity can act as a useful inductive bias rather than purely a handicap.

Several mechanisms may underlie these effects. Parameter sharing in GroupedMLP enforces that the same MLP must serve multiple depths, which may act as an implicit regularizer by discouraging layer-specific overfitting and encouraging more generally useful features. In both GroupedMLP and ShallowMLP, the reduced-capacity networks inherit initialization statistics from a larger model: GroupedMLP through shared parameters derived from the full-capacity ViT-B/16 and ShallowMLP by slicing a wider MLP initialized at the larger width. This suggests that the initialization scale and structure of the underlying large model may be important for optimization, potentially more so than the eventual parameter count in our setting. In addition, shared MLP parameters in GroupedMLP receive gradients from multiple layers, potentially providing richer learning signals, and the empirical effectiveness of the $1/\sqrt{2}$ scaling factor indicates that balancing these contributions is important. The pronounced reduction in peak-to-final accuracy degradation observed in both architectures hints at different loss landscapes, potentially corresponding to flatter minima, but confirming such interpretations will require more direct theoretical and empirical analysis.

\subsection{Limitations and Future Directions}

Our study is intentionally narrow in scope. All experiments are conducted on ViT-B/16 for ImageNet-1K classification under a single training recipe, and we evaluate only two specific parameter-reduction designs. Whether similar gains appear across other model sizes (e.g., ViT-Tiny to ViT-Huge), alternative transformer architectures (such as DeiT or Swin), or other domains (language, multimodal models) remains an open question. We also report results over a limited number of random seeds; although the observed improvements are consistent and statistically significant under our protocol, broader validation across seeds, codebases, and hardware would further strengthen the conclusions. Moreover, we do not analyze downstream transfer, detection, or segmentation tasks, where the inductive biases introduced by parameter sharing or width reduction could help or hinder performance in different ways.

These limitations point naturally to several directions for future work. A first step is to perform systematic scaling studies that vary both model size and the degree of parameter reduction to understand how redundancy and optimal reduction ratios depend on capacity. Exploring richer sharing patterns (beyond adjacent pairs), different group sizes, and alternative scaling rules for shared weights could yield more effective or more flexible designs. Extending the study to other architectures and tasks, including convolutional networks, hybrid models, and language transformers, as well as fine-tuning and transfer-learning setups, would clarify whether the phenomena we observe are specific to vision transformers or reflect broader principles of neural network optimization. Finally, theoretical and empirical analyses of the resulting loss landscapes, gradient dynamics, and information flow could help turn these empirical observations into principled guidelines for designing more efficient and stable transformer architectures.

\section{Conclusion}

We have shown that carefully structured parameter reductions in the MLP blocks of a standard ViT-B/16 can yield modest but consistent accuracy improvements on ImageNet-1K while substantially enhancing training stability. Both of our variants, GroupedMLP, which shares MLP parameters across neighboring blocks, and ShallowMLP, which reduces MLP width, use only 67.3\% of the original parameters yet slightly outperform the full-capacity baseline and exhibit approximately an order-of-magnitude reduction in peak-to-final accuracy degradation. Moreover, the two designs offer complementary practical trade-offs: GroupedMLP preserves the baseline compute cost while reducing memory footprint, and ShallowMLP achieves a 38\% increase in inference throughput by reducing both parameters and FLOPs.

These findings suggest that, for ViT-B/16 on ImageNet-1K under our training setup, there is substantial redundancy in the MLPs and that constraining MLP capacity can serve as a useful inductive bias rather than merely a compression technique. By demonstrating that smaller, more structured MLPs can be both more efficient and better behaved than their overparameterized counterparts, our results add to a growing body of evidence that the organization of capacity in deep networks can matter as much as the total parameter count. We hope this work will motivate further studies that treat parameter reduction and sharing not only as tools for shrinking models, but also as principled mechanisms for improving optimization, robustness, and generalization in transformer architectures.

\bibliographystyle{iclr2026_conference}
\bibliography{references}

@article{ILSVRC15,
Author = {Olga Russakovsky and Jia Deng and Hao Su and Jonathan Krause and Sanjeev Satheesh and Sean Ma and Zhiheng Huang and Andrej Karpathy and Aditya Khosla and Michael Bernstein and Alexander C. Berg and Li Fei-Fei},
Title = {{ImageNet Large Scale Visual Recognition Challenge}},
Year = {2015},
journal   = {International Journal of Computer Vision (IJCV)},
doi = {10.1007/s11263-015-0816-y},
volume={115},
number={3},
pages={211-252}
}

@article{dosovitskiy2020image,
  title={An image is worth 16x16 words: Transformers for image recognition at scale},
  author={Dosovitskiy, Alexey},
  journal={arXiv preprint arXiv:2010.11929},
  year={2020}
}

@article{lan2019albert,
  title={Albert: A lite bert for self-supervised learning of language representations},
  author={Lan, Zhenzhong and Chen, Mingda and Goodman, Sebastian and Gimpel, Kevin and Sharma, Piyush and Soricut, Radu},
  journal={arXiv preprint arXiv:1909.11942},
  year={2019}
}

@article{dehghani2018universal,
  title={Universal transformers},
  author={Dehghani, Mostafa and Gouws, Stephan and Vinyals, Oriol and Uszkoreit, Jakob and Kaiser, {\L}ukasz},
  journal={arXiv preprint arXiv:1807.03819},
  year={2018}
}

@article{frankle2018lottery,
  title={The lottery ticket hypothesis: Finding sparse, trainable neural networks},
  author={Frankle, Jonathan and Carbin, Michael},
  journal={arXiv preprint arXiv:1803.03635},
  year={2018}
}

@inproceedings{liu2021swin,
  title={Swin transformer: Hierarchical vision transformer using shifted windows},
  author={Liu, Ze and Lin, Yutong and Cao, Yue and Hu, Han and Wei, Yixuan and Zhang, Zheng and Lin, Stephen and Guo, Baining},
  booktitle={Proceedings of the IEEE/CVF international conference on computer vision},
  pages={10012--10022},
  year={2021}
}

@inproceedings{rangwani2024deit,
  title={Deit-lt: Distillation strikes back for vision transformer training on long-tailed datasets},
  author={Rangwani, Harsh and Mondal, Pradipto and Mishra, Mayank and Asokan, Ashish Ramayee and Babu, R Venkatesh},
  booktitle={Proceedings of the IEEE/CVF Conference on Computer Vision and Pattern Recognition},
  pages={23396--23406},
  year={2024}
}

@article{nakkiran2021deep,
  title={Deep double descent: Where bigger models and more data hurt},
  author={Nakkiran, Preetum and Kaplun, Gal and Bansal, Yamini and Yang, Tristan and Barak, Boaz and Sutskever, Ilya},
  journal={Journal of Statistical Mechanics: Theory and Experiment},
  volume={2021},
  number={12},
  pages={124003},
  year={2021},
  publisher={IOP Publishing}
}

@article{hoffmann2022training,
  title={Training compute-optimal large language models},
  author={Hoffmann, Jordan and Borgeaud, Sebastian and Mensch, Arthur and Buchatskaya, Elena and Cai, Trevor and Rutherford, Eliza and Casas, Diego de Las and Hendricks, Lisa Anne and Welbl, Johannes and Clark, Aidan and others},
  journal={arXiv preprint arXiv:2203.15556},
  year={2022}
}

@article{foret2020sharpness,
  title={Sharpness-aware minimization for efficiently improving generalization},
  author={Foret, Pierre and Kleiner, Ariel and Mobahi, Hossein and Neyshabur, Behnam},
  journal={arXiv preprint arXiv:2010.01412},
  year={2020}
}

@article{cohen2021gradient,
  title={Gradient descent on neural networks typically occurs at the edge of stability},
  author={Cohen, Jeremy M and Kaur, Simran and Li, Yuanzhi and Kolter, J Zico and Talwalkar, Ameet},
  journal={arXiv preprint arXiv:2103.00065},
  year={2021}
}

@article{belkin2019reconciling,
  title={Reconciling modern machine-learning practice and the classical bias--variance trade-off},
  author={Belkin, Mikhail and Hsu, Daniel and Ma, Siyuan and Mandal, Soumik},
  journal={Proceedings of the National Academy of Sciences},
  volume={116},
  number={32},
  pages={15849--15854},
  year={2019},
  publisher={National Academy of Sciences}
}

@article{rao2021dynamicvit,
  title={Dynamicvit: Efficient vision transformers with dynamic token sparsification},
  author={Rao, Yongming and Zhao, Wenliang and Liu, Benlin and Lu, Jiwen and Zhou, Jie and Hsieh, Cho-Jui},
  journal={Advances in neural information processing systems},
  volume={34},
  pages={13937--13949},
  year={2021}
}

@article{liang2022not,
  title={Not all patches are what you need: Expediting vision transformers via token reorganizations},
  author={Liang, Youwei and Ge, Chongjian and Tong, Zhan and Song, Yibing and Wang, Jue and Xie, Pengtao},
  journal={arXiv preprint arXiv:2202.07800},
  year={2022}
}

@inproceedings{bhojanapalli2021understanding,
  title={Understanding robustness of transformers for image classification},
  author={Bhojanapalli, Srinadh and Chakrabarti, Ayan and Glasner, Daniel and Li, Daliang and Unterthiner, Thomas and Veit, Andreas},
  booktitle={Proceedings of the IEEE/CVF international conference on computer vision},
  pages={10231--10241},
  year={2021}
}

@article{lin2023understanding,
  title={Understanding parameter sharing in transformers},
  author={Lin, Ye and Wang, Mingxuan and Zhang, Zhexi and Wang, Xiaohui and Xiao, Tong and Zhu, Jingbo},
  journal={arXiv preprint arXiv:2306.09380},
  year={2023}
}

@article{chen2021chasing,
  title={Chasing sparsity in vision transformers: An end-to-end exploration},
  author={Chen, Tianlong and Cheng, Yu and Gan, Zhe and Yuan, Lu and Zhang, Lei and Wang, Zhangyang},
  journal={Advances in Neural Information Processing Systems},
  volume={34},
  pages={19974--19988},
  year={2021}
}

@article{zhu2021vision,
  title={Vision transformer pruning},
  author={Zhu, Mingjian and Tang, Yehui and Han, Kai},
  journal={arXiv preprint arXiv:2104.08500},
  year={2021}
}

@article{yu2022unified,
  title={Unified visual transformer compression},
  author={Yu, Shixing and Chen, Tianlong and Shen, Jiayi and Yuan, Huan and Tan, Jianchao and Yang, Sen and Liu, Ji and Wang, Zhangyang},
  journal={arXiv preprint arXiv:2203.08243},
  year={2022}
}

@inproceedings{dehghani2023scaling,
  title={Scaling vision transformers to 22 billion parameters},
  author={Dehghani, Mostafa and Djolonga, Josip and Mustafa, Basil and Padlewski, Piotr and Heek, Jonathan and Gilmer, Justin and Steiner, Andreas Peter and Caron, Mathilde and Geirhos, Robert and Alabdulmohsin, Ibrahim and others},
  booktitle={International conference on machine learning},
  pages={7480--7512},
  year={2023},
  organization={PMLR}
}

@misc{rw2019timm,
  author = {Ross Wightman},
  title = {PyTorch Image Models},
  year = {2019},
  publisher = {GitHub},
  journal = {GitHub repository},
  doi = {10.5281/zenodo.4414861},
  howpublished = {\url{https://github.com/rwightman/pytorch-image-models}}
}

\end{document}